\documentclass[lettersize,journal]{IEEEtran}
\IEEEoverridecommandlockouts

\usepackage{setspace}
\usepackage{cite}
\usepackage{natbib}
\usepackage{amsthm}
\usepackage{subcaption}
\usepackage{lmodern} 
\usepackage{amsmath,amssymb}
\usepackage[type1]{libertine}
\usepackage[T1]{fontenc} 
\usepackage[libertine]{newtxmath}
\usepackage{float}
\usepackage{algorithm}
\usepackage{xcolor}

\DeclareSymbolFont{largesymbols}{OMX}{cmex}{m}{n}
\DeclareMathSymbol{\intop}{\mathop}{largesymbols}{"52}
\DeclareMathSymbol{\sumop}{\mathop}{largesymbols}{"50}
\DeclareMathSymbol{\sqrtop}{\mathop}{largesymbols}{"70}
\usepackage{bm}
\allowdisplaybreaks

\usepackage{algorithmic}
\usepackage{graphicx}
\usepackage{textcomp}
\usepackage[hidelinks, colorlinks=true, linkcolor=blue, urlcolor=blue, citecolor=blue]{hyperref}
\usepackage{cleveref}
\usepackage{commath}
\usepackage{units}
\usepackage{optidef}
\usepackage{orcidlink}
\usepackage{lipsum}
\usepackage{stfloats}

\begin{document}


\IEEEaftertitletext{\vspace{-1.5\baselineskip}}

\title{\LARGE{Meta-Reinforcement Learning for Fast and Data-Efficient Spectrum Allocation in Dynamic Wireless Networks}}

\author{%
  Oluwaseyi~Giwa\orcidlink{0009-0001-5771-7446},
  Tobi~Awodunmila\orcidlink{0009-0001-5637-5827},
  Muhammad~Ahmed~Mohsin\orcidlink{0009-0005-2766-0345},
  Ahsan~Bilal\orcidlink{0009-0002-7044-9316},
  Muhammad~Ali~Jamshed\orcidlink{0000-0002-2141-9025}
  \thanks{Oluwaseyi Giwa and Tobi Awodunmila are with the African Institute for Mathematical Sciences, Muizenberg, 7945, South Africa (e-mail: \{oluwaseyi, tobi\}@aims.ac.za).\newline
  Muhammad Ahmed Mohsin is with Stanford University, Stanford, CA 94305, USA (e-mail: muahmed@stanford.edu).\newline
  Ahsan Bilal is with the School of Computer Science, University of Oklahoma, Norman, OK 73019, USA (e-mail: ahsan.bilal-1@ou.edu).\newline
  Muhammad Ali Jamshed is with the College of Science and Engineering, University of Glasgow, Glasgow G12 8QQ, UK (e-mail: muhammadali.jamshed@glasgow.ac.uk).\newline
  codebase: \url{https://github.com/OluwaseyiWater/meta_learnedRL}.
  }%
}

\makeatletter
\patchcmd{\@maketitle}
{\addvspace{0.5\baselineskip}\egroup}
{\addvspace{0\baselineskip}\egroup}
{}
{}
\makeatother

\maketitle
\begin{abstract}
The dynamic allocation of spectrum in 5G / 6G networks is critical to efficient resource utilization. However, applying traditional deep reinforcement learning (DRL) is often infeasible due to its immense sample complexity and the safety risks associated with unguided exploration, which can cause severe network interference. To address these challenges, we propose a meta-learning framework that enables agents to learn a robust initial policy and rapidly adapt to new wireless scenarios with minimal data. We implement three meta-learning architectures\textemdash model-agnostic meta-learning (MAML), recurrent neural network (RNN), and an attention-enhanced RNN\textemdash and evaluate them against a non-meta-learning DRL algorithm, proximal policy optimization (PPO) baseline, in a simulated dynamic integrated access/backhaul (IAB) environment. Our results show a clear performance gap. The attention-based meta-learning agent reaches a peak mean network throughput of \(\approx 48\) Mbps, while the PPO baseline decreased drastically to \(10\) Mbps.
Furthermore, our method reduces SINR and latency violations by more than \(50\%\) compared to PPO. It also shows quick adaptation, with a fairness index \(\geq 0.7\), showing better resource allocation. This work proves that meta-learning is a very effective and safer option for intelligent control in complex wireless systems.
\end{abstract}


\section{Introduction}\label{introduction}
\IEEEPARstart{T}{he} emergence of fifth generation (5G) and sixth generation (6G) networks and integrated access/backhaul (IAB) architectures creates highly dynamic wireless environments where spectrum resources must be allocated in real-time to meet fluctuating user demands and interference conditions \citep{bhattacharya2022, fang2023}. Deep reinforcement learning (DRL) is a powerful paradigm for this complex decision-making process \citep{sutton1998}. However, its primary drawback is high sample complexity; DRL agents often require millions of interactions to converge \citep{mnih2015}. In a wireless network, this translates to an unacceptably long period of suboptimal performance, leading to dropped calls, high latency, and inefficient resource utilization during the extensive training phase.

Unguided DRL exploration endangers wireless safety: a high-power probe on an occupied band can interfere with nearby cells, breach SLAs, and destabilize the network. While generic safe RL frameworks exist \citep{garcia2015}, they often rely on adding penalties to the reward function. This approach can be inefficient, as the agent must first experience and be penalized for numerous unsafe actions before learning to avoid them. More formal methods like Constrained Markov Decision Processes (CMDP) \citep{altman1999} exist, but do not inherently solve the sample complexity issue.

Proactive constraints like control barrier functions (CBFs) can balance safety and performance in robotics \citep{jabbari2024}, but adapting CBFs to wireless networks’ stochastic, high-dimensional states is not straightforward. Other approaches, like constraint-aware policy optimization, have been proposed \citep{dai2024}, but can be sensitive to the inherent noise and rapidly changing channel dynamics in wireless environments, hindering their reliability. While addressing safety, these methods do not fundamentally reduce the millions of samples required for the agent to learn the underlying network dynamics.

To address both sample inefficiency and safety, we propose a meta-learning framework. The core benefit of meta-learning is its ability to ``learn to learn," improving sample efficiency and generalization across diverse environments \citep{kosaraju2021}. This paradigm has gained momentum in various domains, including multi-agent systems and large language models \citep{bilal2025meta}. For RL, model-agnostic meta-learning (MAML) provides a powerful approach by optimizing for a shared parameter initialization that can be rapidly fine-tuned via gradient descent \citep{finn2017}. We extend this concept by exploring three architectures: a standard MAML implementation, a variant of the recurrent neural network (RNN) to capture temporal dependencies \citep{jacob2023}, and an advanced RNN with a self-attention mechanism designed to better model the complex interplay of states in the wireless network. We evaluate these against proximal policy optimization (PPO), a non-meta-learning baseline to demonstrate their effectiveness.
\section{System Model and Problem Formulation}\label{system-model}
The challenge of dynamic spectrum allocation in 5G/6G networks can be formulated as a CMDP. We aim to find an optimal policy (\(\pi\)) that maximizes network utility while adhering to critical safety and Quality-of-Service (QoS) constraints. The CMDP is defined by the tuple (\(\mathcal{S}, \mathcal{A}, p, R, \mathcal{C}, \gamma\)), where each component is modeled to represent the dynamic wireless environment.
\subsection{State and Action Spaces}
We define the state of the wireless system at time step \(t\) as a tuple \(s_{t}\)
\begin{equation}
    s_t = \left(C_{t}, \; I_{t}, \; Q_{t}, \; A_{t - 1}, \; P_{t - 1}, \; t\right) \in \mathcal{S},
\end{equation}
where \(C_{t} \in \mathbb{R}^{N_{UE} \times N_{BS}}\) is the matrix of the channel gains between the number of base stations (BS), \(N_{BS}\) and the number of user equipments (UE), \(N_{UE}\). These gains are based on a standard log-distance path loss model with additive Gaussian noise. The deterministic component \(L\) is given by:
\begin{equation}
    L = K_{\text{ref}} - 10 \eta \log_{10}d,
\end{equation}
where \(K_{\text{ref}}\) is a reference path loss constant at 1 meter, \(\eta\) is the path loss exponent, and \(d\) is the distance between the transmitter and receiver. The final channel gain matrix is then,
\begin{equation}\label{pathloss}
    C_t = L + \mathcal{N}(0, \sigma^2),
\end{equation}
with \(\sigma^2\) representing the noise variance.

\(I_{t} \in \mathbb{R}^{N_{BS} \times N_{B}}\) is the interference map, detailing inference level on each of the number of frequency bands, \(N_{B}\) at each BS. \(Q_{t} \in \mathbb{R}^{N_{UE} \times 2}\) represents the QoS metrics, containing the current latency and throughput for each UE. \(A_{t - 1}\) and \(P_{t - 1}\) are the spectrum allocation decisions and corresponding power levels from the previous time step, used to model costs associated with switching.

The agent's action at each step, \(a_{t} \in \mathcal{A},\) is a discrete allocation vector \(a_{t} \in \left\{0, 1, \dots, K - 1\right\}^{N_{BS}\times N_{B}}.\) Each element \(a_{t}(i, j)\) corresponds to one of \(K\) discrete power levels for the \(j-\)th band at the \(i-\)th band BS. The environment enforces an action mask based on the current interference map \(I_{t}\) to ensure network safety. For any selected action \(a_{t}\), the environment computes the resulting transmit powers \(P_{t}\) and applies a safety filter to produce the executed action, \(a_{t}^{\text{safe}}\). The power level for each link (\(i, j\)) is set to zero if it violates the maximum permissible interference threshold \(I_{\text{max}}\):
\begin{equation}
    P_t^{\text{safe}}(i,j) = \begin{cases} P_t(i,j) & \text{if } I_t(i,j) < I_{\text{max}} \\ 0 & \text{otherwise} \end{cases}
\end{equation}
where \(P_{t}(i, j)\) is the power corresponding to the selected action \(a_{t}(i, j)\). This hard constraint prevents catastrophically disruptive transmissions.

\subsection{System Dynamics and Channel Model}
The system transitions from state \(s_{t}\) to \(s_{t+1}\) according to the transition probability function \(p(s_{t + 1}|s_{t}, a_{t}).\) This encapsulates the stochastic nature of the wireless environment, including channel fading dynamics, where channel gains evolve according to a first-order autoregressive fading model to simulate temporal correlations:
\begin{equation}\label{fadingmodel}
    C_{t + 1} = \kappa C_{t} + \sqrt{1 - \kappa^{2}}\;\mathcal{N}(0, \sigma_{f}^{2}),
\end{equation}
where \(\kappa\) is the fading coherence factor and \(\sigma_{f}^{2}\) is the variance of the fading process.

Another component of our environment is the Signal-to-Interference-plus-Noise Ratio (SINR). For each active link (BS \(i\) on band \(j\)), the signal power at the intended receiver is a function of the transmit power \(P_{t}(i, j)\), and channel gain \(C_{t}(u, i)\) for the intended user \(u.\) The interference is the sum of powers from all other base stations using the same band, scaled by their respective cross-channel gains. The SINR is therefore:
\begin{equation}
    \text{SINR}_{i,j} = \frac{P_{t}(i, j) \cdot g\left(C_{t}(u, i)\right)}{\sum_{k \neq i} P_{t}(k, j) \cdot g\left(C_{t}(u, k)\right) + N_{0}},
\end{equation}
where \(g(\cdot)\) is a function converting channel gain values to a linear scale, and \(N_{0}\) is the thermal noise power.

The interference map update \(I_{t + 1}\) is another key component of the system dynamics that is updated based on the actions taken. The interference experienced at BS \(i'\)s location on band \(j\) is modeled as the sum of signals from all other base stations transmitting on that band:
\begin{equation}
    I_{t + 1}(i, j) = \sum_{k \neq i} P_{t}(k, j) \cdot h(d_{i, k}),
\end{equation}
where \(h(d_{i, k})\) is a path loss function dependent on the distance \(d_{i, k}\) between base stations \(i\) and \(k\). This updated map is used for the safety masking in the next step. Finally, QoS metrics \(Q_{t}\) are updated based on the power levels selected by action \(a_{t}\).

\subsection{Reward Formulation}
The objective is to learn a policy (\(\pi(a_{t}|s_{t})\)) that maximizes the expected discounted cumulative reward. The reward function \(R(s_{t}, a_{t})\) is designed to balance multiple competing objectives:
\begin{multline}\label{rewardfunc}
 r_t = \underbrace{\omega_1 \, T(s_t, a_t)}_{\text{Throughput Reward}} + \underbrace{\omega_2 \, F(s_t, a_t)}_{\text{Fairness Bonus}} - \underbrace{\omega_3 \, \text{Cost}(s_t, a_t)}_{\text{Power and Switching Cost}} - \\ \underbrace{\omega_4 \, \text{Penalty}(s_t, a_t)}_{\text{Safety Penalty}}
 \end{multline}
 where \(T(s_{t}, a_{t})\) is the total network throughput for all active BS-band pairs, calculated using the Shannon-Hartley theorem.
 \begin{equation}\label{throughput}
     T(s_{t}, a_{t}) = \sum_{i = 1}^{N_{BS}} \sum_{j = 1}^{N_{B}} B\log_{2}(1 + \text{SINR}_{i,j})\; \cdot \; \mathbb{I}(P_{i,j} > 0), 
 \end{equation}
where \(B\) is the channel bandwidth, \(\text{SINR}_{i,j}\) is the SINR for the \(i, j-\)th link, and \(\mathbb{I}(P_{ij} > 0)\) is an indicator function that is 1 if the power \(P_{i, j}\) or that link is non-zero, and 0 otherwise. \(F(s_{t}, a_{t})\) is the fairness parameter which ensures equitable resource distribution among active links; we used Jain's Fairness index. It is calculated over the throughputs \(x_{k}\) of the \(M\) active links (where \(x_{k} = B \log_{2}(1 + \text{SINR}_{k})\)):
\begin{equation}\label{fairness-eqn}
    F(s_{t}, a_{t}) = \frac{\left(\sum_{k = 1}^{M} x_{k}\right)^{2}}{M\sum_{k = 1}^{M} x_{k}^{2}},
\end{equation}
This index ranges from \(\nicefrac{1}{M}\) (worst case) to 1 (perfect fairness, where all active links have the same throughput). \(\text{Cost}(a_{t})\) is the cost term that penalizes total power consumption and the magnitude of changes in power allocation between time steps, which can be disruptive to the hardware. It is defined as:
\begin{equation}
    \text{Cost}(s_{t}, a_{t}) = c_{p} \sum_{i, j} P_{t}(i, j) + c_{s} \sum_{i, j}\left| P_{t}(i, j) - P_{t - 1}(i, j)\right|,
\end{equation}
where \(P_{t}(i, j)\) is the power on the current step, \(P_{t - 1}(i, j)\) is the power from the previous step, and \(c_{p}\) and \(c_{s}\) are the cost coefficients for power and switching, respectively. \(\text{Penalty}\) term incorporates costs for violating critical operational constraints. This includes failing to meet a minimum SINR threshold (\(\text{SINR}_{\text{min}}\)) on active links and exceeding a maximum user latency threshold (\(L_{\text{max}}\)):
\begin{multline}\label{penalty}
    P(s_{t}, a_{t}) = p_{\text{sinr}}\sum_{i, j} \mathbb{I}(\text{SINR}_{i,j} < \text{SINR}_{\text{min}})\; + \\p_{\text{lat}}\sum_{k = 1}^{N_{UE}}\mathbb{I}(Q_{t, k}^{\text{(lat)}} > L_{\text{max}}),
\end{multline}
where \(\mathbb{I}(\cdot)\) is the indicator function, \(p_{\text{sinr}}\) and \(p_{\text{lat}}\) are penalty coefficients. The weights, \(\omega_{n}, \quad \text{for } n \in [1, \dots, 4]\) in Eq.~\ref{rewardfunc} balances the trade-offs between these four objectives. \(c_{p}, \; c_{s}\; p_{\text{sinr}}, \; p_{\text{lat}}\) are hyperparameters that are absorbed into the weights, \(\omega_{3} \; \text{and} \; \omega_{4}\).

\section{Proposed Framework}\label{framework}
\begin{figure*}
    \centering
    \includegraphics[width=0.9\linewidth]{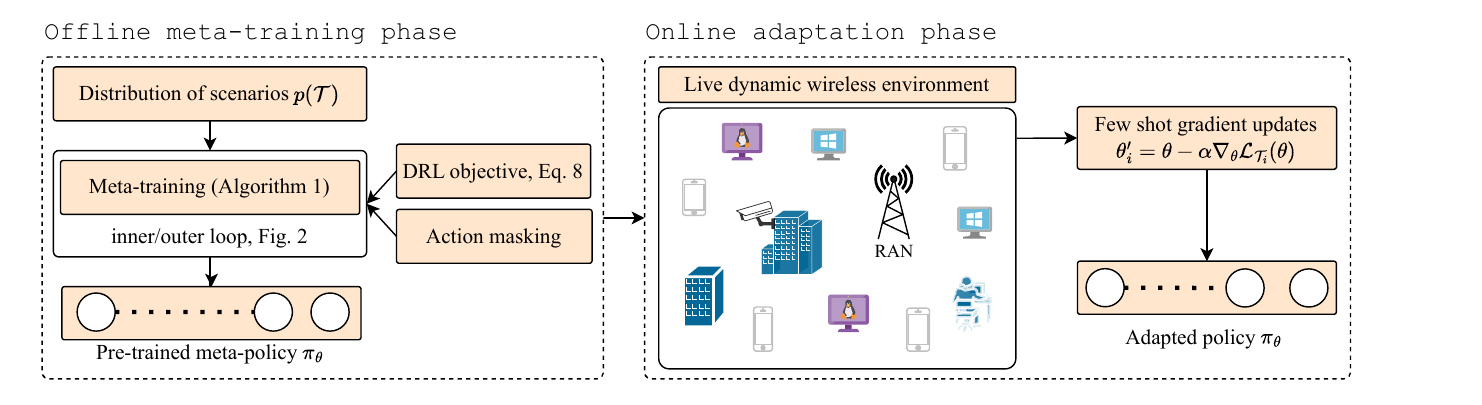}
    \caption{Overview of the proposed framework, divided into two distinct phases. In the Offline Meta-Training Phase (left), the agent learns from a diverse distribution of simulated scenarios. It leverages a DRL objective (Eq.~\ref{rewardfunc}) and safety constraints to produce a single, robust pre-trained meta-policy (\(\pi_\theta\)) that is optimized for rapid adaptation (see Fig.~\ref{meta-rl} for process detail). In the Online Adaptation Phase (right), this pre-trained policy is deployed into a new, live wireless environment. It initializes the local agent and performs a few-shot fine-tuning update to generate a specialized, high-performance policy (\(\pi_{\theta'}\)) tailored to the specific, real-time conditions.}
    \label{drl-meta}
\end{figure*}

\begin{figure*}[t]
    \centering
    \includegraphics[width=0.9\linewidth]{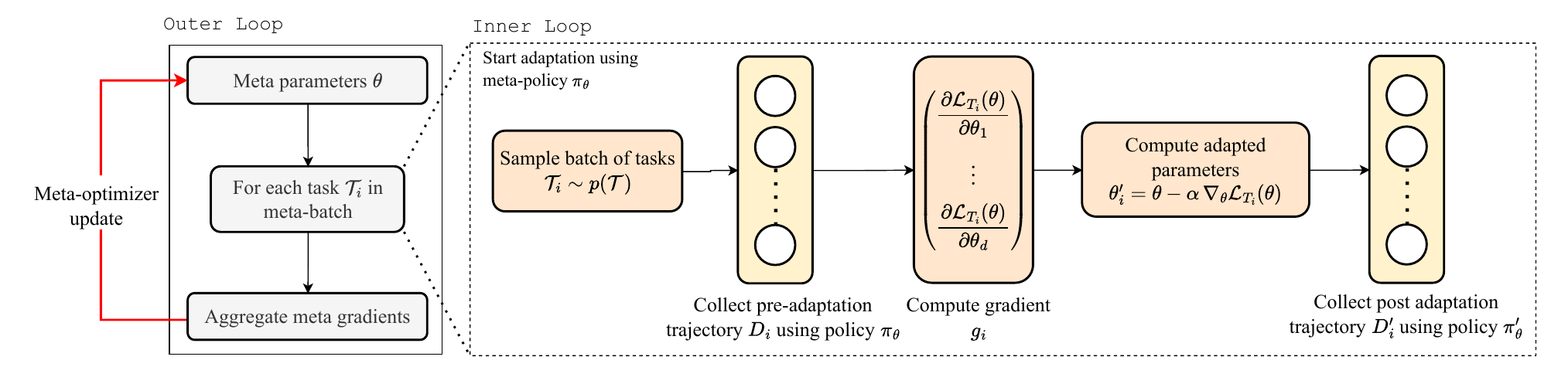}
    \caption{An overview of the two-level optimization process in our meta-learning framework. The outer loop updates the shared meta parameters \(\theta\) by aggregating gradients from a batch of tasks. For each task, the inner loop performs a fast adaptation by taking the current meta parameters \(\theta\), collecting a support set of trajectories, and computing adapted parameters \(\theta'_{i}\) via one or more gradient steps. The performance of this adapted policy \(\pi_{\theta'_i}\) on a query set provides the learning signal for the outer loop, enabling the agent to ``learn how to adapt." }
    \label{meta-rl}
\end{figure*}

To overcome the challenges of sample inefficiency and unsafe exploration, our proposed framework leverages meta-learning to train agents that can rapidly adapt to novel wireless network conditions. As illustrated in Fig.~\ref{drl-meta}, the overall architecture consists of two sequential phases: an offline meta-training phase, where the agent learns a robust initial policy across a wide distribution of simulated scenarios, and an online adaptation phase, where the pre-trained agent is deployed and quickly fine-tunes its policy to the specific, live environment.
The core of the offline meta-training phase is a two-level optimization process, depicted in detail in Fig.~\ref{meta-rl}. This approach is ``model-agnostic" in the sense that it can be applied to various policy network architectures. We implement this framework with three distinct architectures: a standard feed-forward network, MAML, an RNN, and an advanced RNN with a self-attention mechanism.
For any of these architectures, the meta-learning objective is to find a single set of initial parameters \(\theta\) that serves as an effective starting point for rapid adaptation. In the inner loop of the framework, a task-specific policy is formed by taking a few gradient steps on data from a new task \(\mathcal{T}_i\):
\begin{equation}\label{metaobj}
    \theta'_{i} = \theta - \alpha\nabla_{\theta}\mathcal{L}_{\mathcal{T}_{i}}(\theta)
\end{equation}
The outer loop then updates the meta parameters \(\theta\) by minimizing the expected loss of the adapted policies \(\pi_{\theta'_i}\) over the distribution of all tasks:
\begin{equation}\label{metaobj2}
   \underset{\theta}{\text{min}} \sum_{\mathcal{T}_{i} \sim p(\mathcal{T})} \mathcal{L}_{\mathcal{T}_{i}}(\theta') = \underset{\theta}{\text{min}} \sum_{\mathcal{T}_{i} \sim p(\mathcal{T})} \mathcal{L}_{\mathcal{T}_{i}} \left(\theta - \alpha \nabla_{\theta}\mathcal{L}_{\mathcal{T}_{i}}(\theta)\right),
\end{equation}
This process is formalized in Algorithm~\ref{alg:meta_training}, enabling the agent to acquire a broad, generalizable understanding of wireless dynamics, ensuring both fast and safe adaptation during the online phase. Safety is further enhanced by an environment-level action mask that prevents transmissions violating a maximum interference threshold, \(I_{max}\), and by reward penalties for softer QoS constraint violations.

\begin{algorithm}[t]
\caption{Meta-Training Framework for Adaptive Policies}
\label{alg:meta_training}
\begin{algorithmic}[1]
\STATE Input: Task distribution \(p(\mathcal{T})\), inner learning rate \(\alpha\), meta-learning rate \(\beta\)
\STATE Initialize meta parameters \(\theta\) for a chosen policy/value architecture (MAML, RNN, or RNN+Attention).
\STATE Initialize meta-optimizer state \(\phi\).
\FOR{iteration = \(1, 2, \dots, N_{meta\_iters}\)}
\STATE Sample a meta-batch of \(B\) tasks \(\mathcal{T}_1, \mathcal{T}_2, \dots, \mathcal{T}_B\) from the task distribution \(p(\mathcal{T})\)
\FOR{each task \(\mathcal{T}_i\) in the meta-batch} 
    \STATE Collect trajectory \(\mathcal{D}_i = \{ (s_t, a_t, r_t, s_{t+1}) \}\)
    \STATE Compute the task-specific loss \(\mathcal{L}_{\mathcal{T}_i}(\theta)\) on \(\mathcal{D}_i\).
    \STATE Compute the inner-loop gradient: \(g_i \leftarrow \nabla_\theta \mathcal{L}_{\mathcal{T}_i}(\theta)\)
    \STATE Adapt parameters for task: \(\theta'_i \leftarrow \theta - \alpha g_i\) (Eq.~\ref{metaobj})
    \STATE Collect trajectory \(\mathcal{D}'_i\)
    \STATE Compute the post-adaptation loss: \(\mathcal{L}_{\mathcal{T}_i}(\theta'_i)\) on \(\mathcal{D}'_i\).
\ENDFOR
\STATE Define the meta-objective over the batch:
\STATE \(\mathcal{L}_{\text{meta}}(\theta) = \frac{1}{B} \sum_{i=1}^{B} \mathcal{L}_{\mathcal{T}_i}(\theta'_i) = \frac{1}{B} \sum_{i=1}^{B} \mathcal{L}_{\mathcal{T}_i}(\theta - \alpha \nabla_\theta \mathcal{L}_{\mathcal{T}_i}(\theta))\) (Eq.\ref{metaobj2})
\STATE Compute aggregated meta-gradient: \(\nabla_{\text{meta}} \leftarrow \nabla_\theta \mathcal{L}_{\text{meta}}(\theta)\)
\STATE Update the global meta parameters:
\STATE \((\theta, \phi) \leftarrow \text{Meta-optimizer update}(\nabla_{\text{meta}}, \phi, \theta)\)
\ENDFOR
\STATE \textbf{return} Optimized meta parameters \(\theta\)
\end{algorithmic}
\end{algorithm}
Here, \(\alpha\) and \(\beta\) denote the inner and meta-learning rates, respectively. The loss \(\mathcal{L}_{\mathcal{T}_{i}}\) is calculated using the DRL objective, which includes the aforementioned reward function Eq.~\ref{rewardfunc} and incorporates the safety penalty to discourage unsafe actions.

\section{Experimental Setup}\label{simul}
\subsection{Simulation Environment}
 To evaluate our proposed framework, we developed a high-fidelity simulation of a dynamic 5G/6G wireless network featuring an Integrated Access and Backhaul (IAB) architecture. The environment is configured with \(N_{BS} = 3, \; N_{UE} = 10, \; N_{B} = 5,\) with agents selecting from \(K = 4\) discrete power levels.
 The system dynamics are stochastic, incorporating key real-world phenomena: channel gains evolve according to a temporal fading model in Eq.~\ref{fadingmodel} with a coherence factor (\(\kappa = 0.9\)), and task variations are introduced by sampling different path loss scenarios (\(L\) in Eq.~\ref{pathloss}) and interference maps \(I_{0}\) at the start of each episode. This setup creates a rich and realistic distribution of tasks, providing a challenging testbed for evaluating the adaptation capabilities of the meta-learning algorithms.

 \subsection{Performance Metrics}
To provide a comprehensive evaluation of the algorithms, we assess their performance based on four key metrics that capture both network efficiency and safety:
\begin{itemize}
    \item[1.] Mean SINR Violations: To specifically evaluate the agent's ability to adhere to safety and QoS constraints, we measure the average number of SINR violations per episode. A violation occurs whenever an active communication link's SINR falls below a predefined minimum threshold, \(\text{SINR}_{\text{min}}\). A lower value for this metric signifies a safer and more reliable policy.
    \item[2.] Mean Network Throughput: To measure the primary objective of network utility, we calculate the total throughput achieved during an episode. This is given by Eq.~\ref{throughput}. We report the average total throughput per episode.
    \item[3.] Mean Latency Violations: This metric evaluates the agent's ability to maintain Quality of Service (QoS) for users. A latency violation is counted for each user whose accumulated service delay exceeds a maximum threshold, \(L_{\text{max}}\), as defined in the penalty term of Eq.~\ref{penalty}.
    \item[4.] Fairness Index: To assess how equitably the agent distributes resources, we employ Jain's Fairness Index, formally defined in Eq.~\ref{fairness-eqn}. The index ranges from a value near zero (highly unfair allocation) to a maximum of 1 (perfect fairness, where all users receive the same throughput).
\end{itemize}
These four metrics allow for a direct comparison of the meta-learning approaches against the non-meta-learning PPO baseline in terms of both their ability to learn an effective control policy and their capacity to operate within critical network constraints.

\section{Results and Discussions}\label{results}
To evaluate the effectiveness of our proposed meta-learning framework, we conducted a comparative analysis of three meta-learning algorithms against a standard, non-meta-learning baseline. The algorithms implemented were MAML (orange dashed lines), RNN (yellow dotted lines), and a more advanced RNN variant (blue solid line), incorporating a self-attention mechanism (RNN + Attention Mechanism). These were compared to PPO (purple dash-dotted lines), a popular DRL algorithm. The evaluation took place over 1200 episodes. It focused on four main performance metrics: the average number of SINR violations, latency violations, network throughput, and the fairness index.

\begin{figure}[t!] 
    \centering 
    \begin{subfigure}[b]{0.49\columnwidth}
        \centering
        \includegraphics[width=\textwidth]{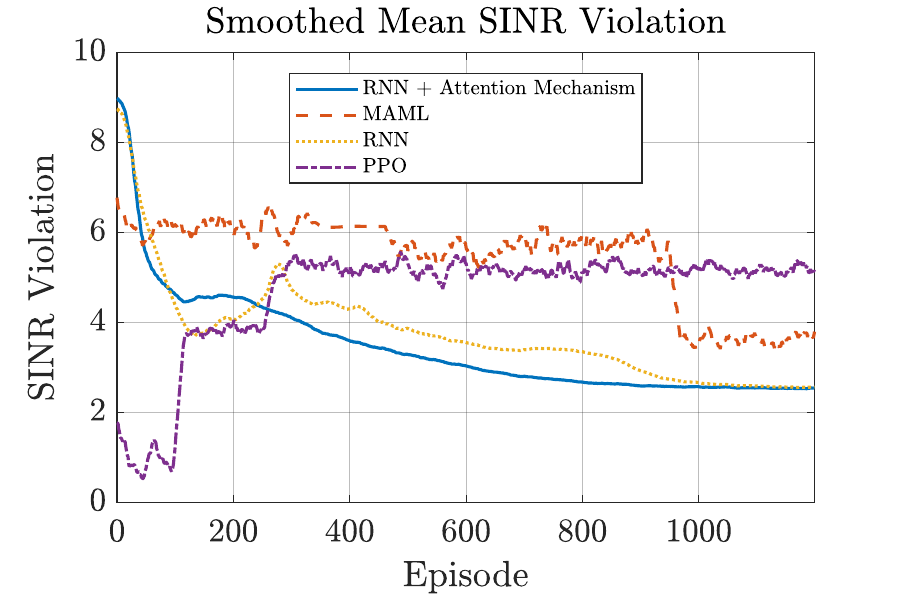}
        \caption{Mean SINR violations.}
        \label{sinr_vio}
    \end{subfigure}
    \hfill 
    \begin{subfigure}[b]{0.49\columnwidth}
        \centering
        \includegraphics[width=\textwidth]{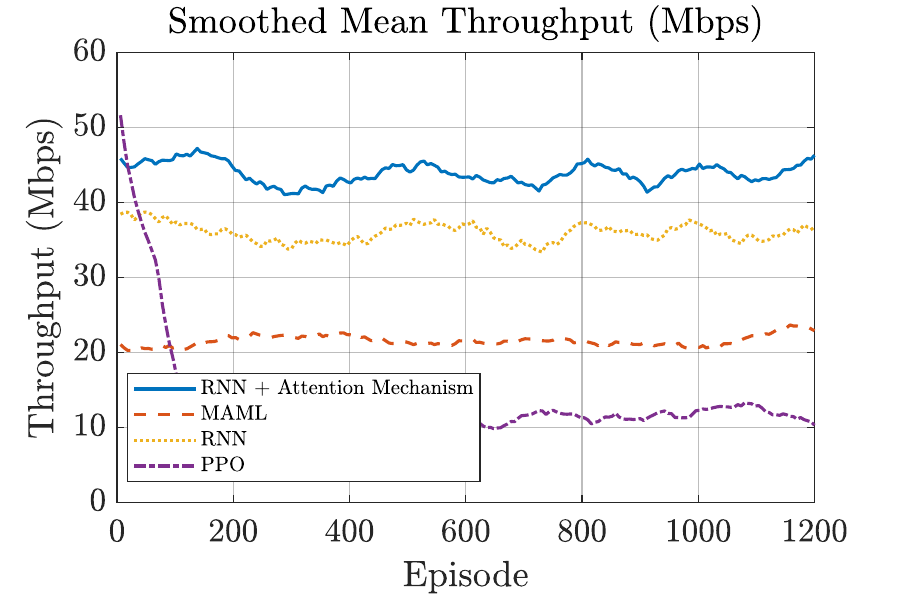}
        \caption{Network throughput (Mbps).}
        \label{network_throughput}
    \end{subfigure}

     \vspace{0.2cm} 
   
    \begin{subfigure}[b]{0.49\columnwidth} 
        \centering
        \includegraphics[width=\textwidth]{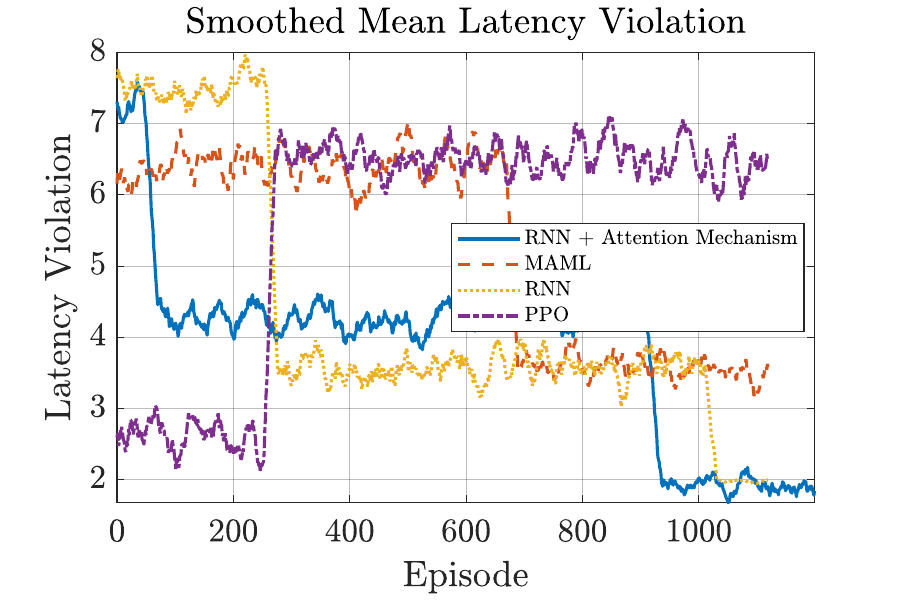}
        \caption{Latency violations.}
        \label{latency_vio}
    \end{subfigure}
    \hfill 
    \begin{subfigure}[b]{0.49\columnwidth} 
        \centering
        \includegraphics[width=\textwidth]{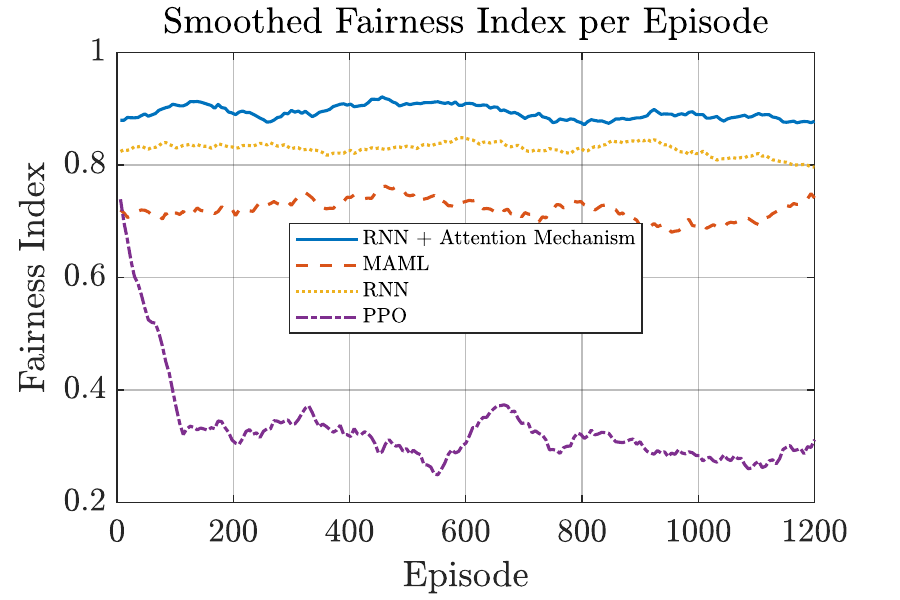}
        \caption{Fairness index.}
        \label{fairness_index}
    \end{subfigure}
    \hfill
    \caption{Comparative performance of the four algorithms on all four performance metrics.}
    \label{fig:overall_performance}
\end{figure}

The results demonstrate a clear performance gap between the meta-learning approaches and the PPO baseline across all metrics. As shown in Fig.~\ref{network_throughput}, the meta-learning agents achieve substantial network throughput, with the RNN + Attention variant reaching the highest peak. Conversely, the PPO agent fails to learn an effective transmission strategy, resulting in a very low throughput for the duration of the training. This indicates that the meta-learning agents are significantly more sample-efficient at discovering policies that maximize network utility.

 This performance advantage extends to safety and Quality of Service. The SINR and latency violation results (Figs.~\ref{sinr_vio},~\ref{latency_vio}) show that while the PPO agent consistently violates constraints, all meta-learning agents quickly learn to operate safely. The recurrent models prove most adept, leveraging their temporal memory to achieve the lowest and most stable violation rates. Furthermore, the meta-learning agents also demonstrate superior resource allocation, achieving a high degree of fairness as measured by the Jain's Fairness Index (Fig.~\ref{fairness_index}), whereas PPO's inability to establish a useful policy results in poor fairness.
 
In summary, the experiments confirm that meta-learning provides a robust framework for developing agents that simultaneously learn to improve network utility and adhere to critical operational rules, far exceeding the capabilities of a standard DRL baseline in this complex, dynamic wireless environment.

\section{Conclusion}\label{conclusion}
Dynamic 5G/6G spectrum use is hampered by traditional DRL's high sample demands and unsafe exploration, leading to prolonged performance drops and interference. In this study, we tackled this problem by proposing a meta-learning framework aimed at developing a strong initial policy that can quickly and safely adjust to new network conditions. We implemented and compared three meta-learning architectures: MAML, RNN, and RNN with an attention mechanism. We tested these against a standard PPO baseline. Our experimental results showed that all meta-learning agents significantly outperformed PPO. They achieved much higher network throughput and better resource allocation, while causing far fewer SINR and latency violations. The agent that combined recurrence and self-attention was the most effective, highlighting the benefit of using advanced architectural designs. Our results indicate that meta-learning offers a practical way to create data-efficient agents that are not only high-performing but also safer. This makes them better suited for complex wireless systems in the real world. Future work will focus on adding more formal safety measures and testing this framework on hardware platforms.

\bibliographystyle{unsrt}
\bibliography{references/references}

\end{document}